\documentclass[10pt,twocolumn,letterpaper]{article}

\usepackage{iccv}
\usepackage{times}
\usepackage{epsfig}
\usepackage{graphicx}
\usepackage{amsmath}
\usepackage{amssymb}
\usepackage{mathtools}
\usepackage{booktabs}
\usepackage{caption}
\usepackage{subcaption}
\usepackage[sort,nocompress]{cite}

\newcommand{\norm}[1]{\left\lVert #1 \right\rVert}

\DeclarePairedDelimiter{\floor}{\lfloor}{\rfloor}

\usepackage[pagebackref=true,breaklinks=true,letterpaper=true,colorlinks,bookmarks=false]{hyperref}

\iccvfinalcopy

\ificcvfinal\fi

\newcommand{\bb}{\mathbf{b}}
\newcommand{\bc}{\mathbf{c}}
\newcommand{\bd}{\mathbf{d}}

\newcommand{\bi}{\mathbf{i}}

\newcommand{\br}{\mathbf{r}}

\newcommand{\bx}{\mathbf{x}}

\newcommand{\cL}{\mathcal{L}}

\newcommand{\figref}[1]{Fig.~\ref{#1}}

\makeatletter
\DeclareRobustCommand\onedot{\futurelet\@let@token\@onedot}
\def\@onedot{\ifx\@let@token.\else.\null\fi\xspace}
\def\eg{e.g\onedot} 
\def\ie{i.e\onedot}

\def\wrt{wrt\onedot}

\def\etal{et~al\onedot}

\makeatother

\newcommand{\boldparagraph}[1]{\vspace{0.2cm}\noindent{\bf #1:}}

\begin{document}
	
\title{KiloNeRF: Speeding up Neural Radiance Fields with Thousands of Tiny MLPs}

\author{Christian Reiser$^{1,2}$ \quad Songyou Peng$^{1,3}$ \quad Yiyi Liao$^{1,2}$ \quad Andreas Geiger$^{1,2}$\\
	$^1$Max Planck Institute for Intelligent Systems, T{\"u}bingen \quad $^2$University of T{\"u}bingen \quad $^3$ETH Zurich\\
	{\tt\small \{firstname.lastname\}@tue.mpg.de}
}

\maketitle

\ificcvfinal\thispagestyle{empty}\fi

\begin{abstract}
NeRF synthesizes novel views of a scene with unprecedented quality by fitting a neural radiance field to RGB images. However, NeRF requires querying a deep Multi-Layer Perceptron (MLP) millions of times, leading to slow rendering times, even on modern GPUs. In this paper, we demonstrate that real-time rendering is possible by utilizing thousands of tiny MLPs instead of one single large MLP. In our setting, each individual MLP only needs to represent parts of the scene, thus smaller and faster-to-evaluate MLPs can be used. By combining this divide-and-conquer strategy with further optimizations, rendering is accelerated by three orders of magnitude compared to the original NeRF model without incurring high storage costs. Further, using teacher-student distillation for training, we show that this speed-up can be achieved without sacrificing visual quality.
\end{abstract}
\section{Introduction}

Novel View Synthesis (NVS) addresses the problem of rendering a scene from unobserved viewpoints, given a number of RGB images and camera poses as input, \eg, for interactive exploration.
Recently, NeRF \cite{mildenhall2020nerf} demonstrated state-of-the-art results on this problem using a neural radiance field representation for representing 3D scenes.
NeRF produces geometrically consistent, high-quality novel views even when faced with challenges like thin structures, semi-transparent objects and reflections. Additionally, NeRF's underlying representation requires only very little storage and thus can be easily streamed to users.

The biggest remaining drawbacks of NeRF are its long training and rendering times. While training can be sped up with a multi-GPU cluster, rendering must happen in real-time on the consumer's device for interactive applications like virtual reality. This motivates us to focus on increasing NeRF's rendering speed in this paper.

\begin{figure}
	\includegraphics[width=\linewidth]{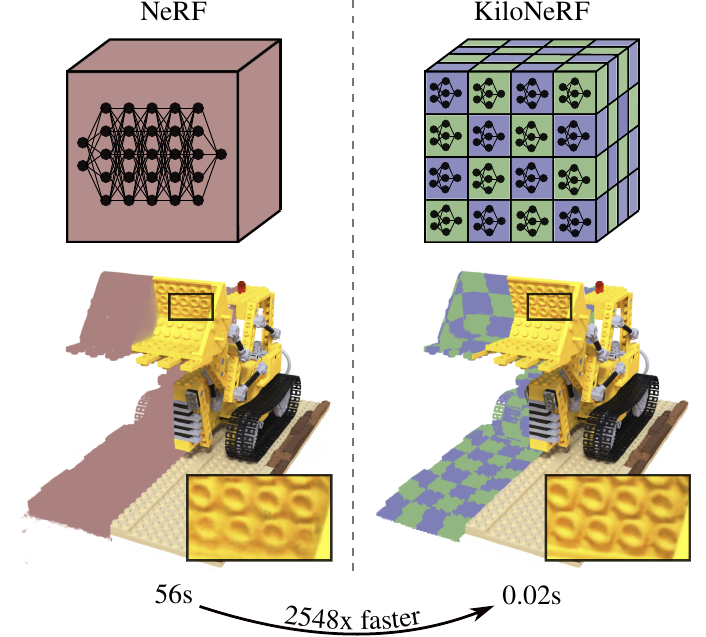}
	\caption{\textbf{KiloNeRF.} Instead of representing the entire scene by a single, high-capacity MLP, we represent the scene by thousands of small MLPs. This allows us to render the scene above 2548x faster without loss in visual quality.}
	\label{fig:teaser}
\end{figure}

NeRF represents the scene's geometry and appearance with a Multi-Layer Perceptron (MLP). During volumetric rendering, this network is sampled hundreds of times for millions of pixels.  As the MLP used in NeRF is relatively deep and wide, this process is very slow. A natural idea is to decrease the depth and number of hidden units per layer in order to speed up the rendering process. However, without any further measures, a reduction in network size leads to an immediate loss in image quality due to the limited capacity for fitting complex scenes. We counteract this by using a large number of independent and small networks, and by letting each network represent only a fraction of the scene.

We find that training our KiloNeRF with thousands of networks, na\"ively from scratch leads to noticeable artifacts. To overcome this problem, we first train a regular NeRF as teacher model. KiloNeRF is then trained such that its outputs (density and color) match those of the teacher model for any position and view direction. Finally, KiloNeRF is fine-tuned on the original training images. Thanks to this three-stage training strategy, our model reaches the same visual fidelity as the original NeRF model, while being able to synthesize novel views three orders of magnitude faster as illustrated in \figref{fig:teaser}.
Crucial for achieving our speedup is an adequate implementation of the concurrent evaluation of many MLPs. Towards this goal, we published our efficient implementation using PyTorch, MAGMA, Thrust and custom CUDA kernels at \href{https://github.com/creiser/kilonerf}{https://github.com/creiser/kilonerf}.

\section{Related Work}

\boldparagraph{Novel View Synthesis}
NVS methods can be categorized according to the representations they use to model the underlying scene geometry.
Mesh-based methods, including classical~\cite{Debevec1996SIGGRAPH,Waechter2014ECCV,Buehler2001SIGGRAPH,Wood2000SIGGRAPH} and learning-based~\cite{Thies2019TOG,riegler2020fvs,riegler2020svs} ones, typically require a preprocessing step, \eg, Structure from Motion (SfM)~\cite{Schoenberger2016CVPRa}, to reconstruct the scene. Similarly, point cloud-based methods rely on SfM or RGB-D sensors to recover geometry~\cite{Aliev2020ECCV}.
In contrast, approaches using multi-plane images (MPIs)~\cite{Mildenhall2019SIGGRAPH,Flynn2019CVPR,Srinivasan2019CVPR,Srinivasan2020CVPR,Zhou2018SIGGRAPH,Tucker2020CVPR} represent the scene as a stack of images or feature maps. While MPI approaches demonstrate photorealistic image synthesis, they only allow for small viewpoint changes during inference.
Another line of methods considers voxel grids as scene representation~\cite{Seitz1999IJCV,Kutulakos2000IJCV,Szeliski1998ICCV,Penner2017SIGGRAPH,Sitzmann2019CVPR,lombardi2019nv} which, however, are typically restricted in terms of their resolution.

The limitations of voxel grids can be alleviated by adopting neural function representations~\cite{Sitzmann2019NIPS,niemeyer2020dvr,Yariv2020NeurIPS,mildenhall2020nerf}. Differentiable Volumetric Rendering (DVR) \cite{niemeyer2020dvr} and Implicit Differentiable Renderer (IDR)~\cite{Yariv2020NeurIPS} adopt surface rendering and therefore rely on pixel-accurate object masks as input during training. Furthermore, they assume solid scenes and cannot handle semi-transparent objects or thin structures well. In contrast, NeRF~\cite{mildenhall2020nerf} uses volumetric rendering \cite{beyer2015volsurv} which enables training without masks and allows for recovering fine structures using alpha blending.

One reason for NeRF's success lies in its representation, which utilizes a parametric function to map 3D coordinates and viewing directions to volumetric densities and color values. Such function representations have a long tradition \cite{savchenko1995func, schoelkopf2005kernel} and have recently resurfaced in geometric computer vision \cite{mescheder2019occnet, park2019deepsdf, chen2018imnet}. These recent methods use deep neural networks as function class and have advanced the state-of-the-art in tasks like 3D reconstruction, shape generation, point cloud completion or relighting \cite{mescheder2019occnet, park2019deepsdf, chen2018imnet, peng2020convocc, oechsle2020surface}.

\boldparagraph{Faster NeRF Rendering}
Neural Sparse Voxel Fields (NSVF) \cite{liu2020nsvf} speed up NeRF's rendering using classical techniques like empty space skipping and early ray termination. Additionally, NSVF's network is conditioned on feature vectors located on a uniform 3D grid to increase the model's capacity without increasing network depth or width.
By applying empty space skipping already during training, NSVF can sample more densely around surfaces at the same computational budget. We also make use of empty space skipping during training and early ray termination during rendering. The key difference is that we use thousands of small networks, while NSVF uses a single feature-conditioned network. Since our networks are only responsible for representing a small region, they require lower capacity compared to NSVF's single network that must represent the entire scene. As a consequence, KiloNeRF renders views two orders of magnitude faster than NSVF.

Concurrently to us, numerous works \cite{rebain2020derf, lindell2020autoint, Neff2021donerf, Garbin2021ARXIV, Yu2021ARXIV, Hedman2021ARXIV} were developed with the purpose of speeding up NeRF.
DeRF \cite{rebain2020derf} also represents the scene by a number of independent networks. However, in DeRF the scene is decomposed into sixteen irregular Voronoi cells. In this paper, we demonstrate that a much simpler strategy of decomposing the scene into thousands of MLPs arranged on a regular 3D grid leads to significantly higher speedups.
AutoInt replaces the need for numerical integration in NeRF by learning a closed form solution of the antiderivative \cite{lindell2020autoint}.
By making use of the fundamental theorem of calculus, a pixel is rendered with a significantly smaller amount of network queries.
DONeRF \cite{Neff2021donerf} demonstrates the feasibility to render a pixel using only 4 samples on the ray. This is achieved by placing samples more closely around the first surface that the ray intersects. Towards this goal, for each ray an additional network is queried, that directly predicts suitable sample locations.
We note that the latter two approaches are orthogonal to KiloNeRF and the combination of KiloNeRF with either of these new techniques is promising. Other works show that real-time rendering can be achieved by converting the neural representation into a discrete one after training \cite{Garbin2021ARXIV, Yu2021ARXIV, Hedman2021ARXIV}. In comparison to KiloNeRF, these approaches consume significantly more GPU memory which might make them less suitable for larger scenes.

\boldparagraph{NeRF Follow-ups}
NeRF++ extends NeRF to unbounded scenes \cite{zhang2020nerfpp} while NeRF-W tackles unstructured photo collections \cite{martinbrualla2021nerfw}. GRAF \cite{schwarz2020graf}, pi-GAN \cite{chan2020pigan} and GIRAFFE \cite{niemeyer2020giraffe} propose generative models of radiance fields. A series of works improve generalization from limited number of training views \cite{trevithick2020grf, yu2020pixelnerf, tancik2020meta, gao2020portrait, rematas2021sharf,Wang2021CVPR,Raj2021ARXIV} while others  \cite{YenChen2020ARXIV,Su2021ARXIV,Wang2021ARXIV} remove the requirement for pose estimation. \cite{boss2020nerd,Guo2020ARXIV,Srinivasan2020ARXIV} enable learning from images with varying light conditions. Finally, ~\cite{li2020flow,xian2020spacetime,du2020neural,peng2020neural,park2020deformable,pumarola2020dnerf,gafni2020dynamic,Wang2020ARXIV,Li2021ARXIV,Ost2021CVPR,tretschk2020nonrigid} extend NeRF to videos.
Many of these methods would benefit from faster rendering and are compatible with KiloNeRF.

\section{Method}

Our method builds upon NeRF \cite{mildenhall2020nerf}, which represents a single scene as a collection of volumetric density and color values. Crucially, in NeRF those densities and colors are not stored in a discrete voxel representation. Instead, NeRF encodes densities and colors at any continuous 3D position in the scene using an MLP.
While NeRF uses a single network to represent the entire scene, we are inspired by \cite{dai_2015_regress} and represent the scene with a large number of independent and small MLPs. More specifically, we subdivide the scene into a 3D grid. Each MLP is tasked to represent the part of the scene that falls within a particular 3D cell. Since each network only represents a small portion of the scene, a low network capacity is sufficient for photo-realistic results as demonstrated by our experiments. 
Additionally, we employ empty space skipping and early ray termination to speed up rendering further. We start with a brief review of the original NeRF model which forms the basis for KiloNeRF.

\subsection{Background}

In NeRF \cite{mildenhall2020nerf}, the scene is represented by a neural network $f_\theta$ with parameters $\theta$. $f_\theta$ takes as input a 3D position $\bx$ and viewing direction $\bd$ and maps them to a color $\bc$ and density $\sigma$. The architecture of $f_\theta$ is chosen such that only the color $\bc$ depends on the viewing direction $\bd$. This allows modeling of view-dependent effects like specularities and reflections while also encouraging a consistent geometry to be learned.
In a deterministic pre-processing step, $\bx$ and $\bd$ are transformed by a positional encoding $\gamma$ which promotes learning of high-frequency details, see \cite{mildenhall2020nerf, tancik2020fourfeat} for details.

A pixel is rendered by shooting a ray from the camera's eye through the pixel's center and evaluating the network $f_\theta$ for $K$ sample points $\bx_1, \ldots, \bx_K$ along the ray. For each sample $\bx_i$, the network outputs a color value $\bc_i$ and a density value $\sigma_i$. In other words, we compute a list of tuples:
\begin{equation}
	(\bc_i, \sigma_i) = f_\theta(\bx_i, \bd) \quad \text{ with } \quad i = 1, 2, \ldots, K
\end{equation}
Here, the viewing direction $\bd$ is computed by normalizing the vector from the camera center to the pixel. Subsequently, the final pixel color $\hat{\bc}$ is calculated by $\alpha$-blending the color values $\bc_1, \ldots, \bc_K$
\begin{align}
	\hat{\bc} &= \sum_{i=1}^{K} T_i \alpha_i \bc_i \label{eq:render}\\
	T_i &= \prod_{j=1}^{i-1} (1 - \alpha_j)\\
	\alpha_i &= 1 - \exp(\sigma_i \delta_i) \label{eq:density_to_alpha}
\end{align}
where $\alpha_i$ determines the alpha value used for blending the color values and the calculation of $\alpha_i$ depends on the distance between adjacent sample points $\delta_j = \norm{\bx_{j+1} - \bx_j}$. Finally, $T_i$ corresponds to the transmittance which accounts for occlusion along the ray. This procedure needs to be repeated for every pixel in the image. Consequently, $W \times H \times K$ network evaluations are required for rendering an image of width $W$ and height $H$.

NeRF is trained by minimizing a photometric loss between rendered images and training images. More specifically, for each parameter update step, we randomly sample a training image and $B$ pixels $\bc_1, \ldots, \bc_B$ inside that image. Subsequently, the corresponding pixels $\hat{\bc}_1, \ldots, \hat{\bc}_B$ are rendered according to Eq.~\ref{eq:render}. The model parameters $\theta$ are optimized by minimizing an $L_2$ reconstruction loss between the rendered pixels and their ground truth:
\begin{equation} \label{eq:photometric_loss}
	\cL = \frac{1}{B} \sum_{b}^{B} \lVert\bc_b - \hat{\bc}_b\rVert_2^2
\end{equation}

\subsection{KiloNeRF}

KiloNeRF assumes knowledge of an axis aligned bounding box (AABB) enclosing the scene. Let $\bb_{min}$ and $\bb_{max}$ be the minimum and maximum bounds of this AABB. We subdivide the scene into a uniform grid of resolution $\br = (r_x, r_y, r_z)$. Each grid cell with 3D index $\bi = (i_x, i_y, i_z)$ corresponds to a tiny MLP network with an independent set of parameters $\theta(\bi)$. The mapping $g$ from position $\textbf{x}$ to index $\bi$ of the tiny network assigned to position $\textbf{x}$ is defined through spatial binning
\begin{equation}
	g(\bx) = \floor{(\bx - \bb_{min}) / ((\bb_{max} - \bb_{min}) /\br)}
\end{equation}
where all operations are element-wise. Querying KiloNeRF at a point $\bx$ and direction $\bd$ involves determining the network $g(\bx)$ responsible for the respective grid cell:
\begin{equation}
	(\bc, \sigma) = f_{\theta(g(\bx))}(\bx, \bd)
\end{equation}

\begin{figure*}
	\begin{center}
		\includegraphics[width=\linewidth]{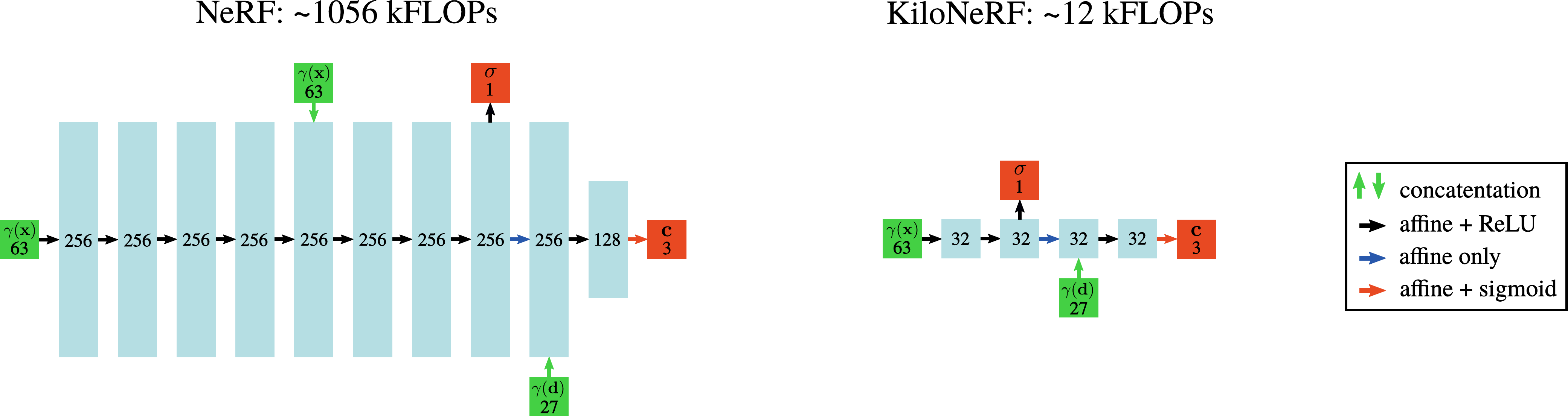}
	\end{center}
	\caption{\textbf{Model Architecture.} KiloNeRF's MLP architecture is a downscaled version of NeRF's architecture. A forward pass through KiloNeRF's network only requires 1/87th of the floating point operations (FLOPs) of the original architecture.}
	\label{fig:arch}
\end{figure*}

\boldparagraph{Network Architecture}
As illustrated in Fig.~\ref{fig:arch}, we use a downscaled version of the fully-connected architecture of NeRF which similarly to the NeRF architecture enforces that the predicted density is independent of the view direction. However, NeRF uses a total of 10 hidden layers where each of the first 9 hidden layers outputs a 256-dimensional feature vector and the last hidden layer outputs a 128-dimensional feature vector. In contrast, we use a much smaller MLP with only 4 hidden layers and 32 hidden units each. Due to the low depth of our network, a skip connection as in the original NeRF architecture is not required. Like in NeRF, we provide the viewing direction as an additional input to the last hidden layer. All affine layers are followed by a ReLU activation with two exceptions: The output layer that computes the RGB color $\bc$ uses a sigmoid activation and no activation is applied to the feature vector that is given as input to the penultimate hidden layer.

\begin{figure}
    \begin{subfigure}{0.48\linewidth}
		\includegraphics[width=\linewidth]{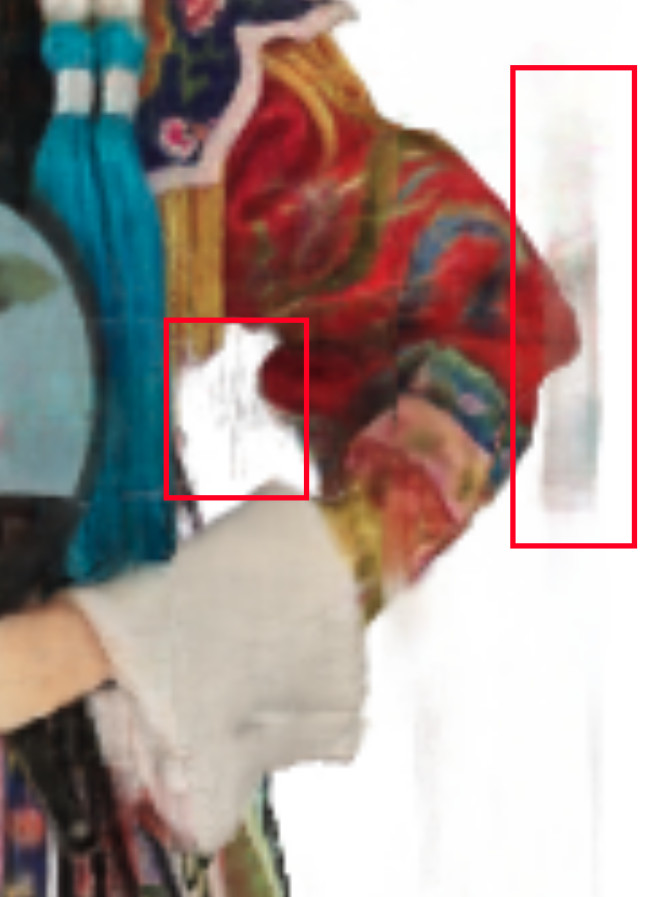}
		\caption{Without Distillation}
		\label{fig:from_scratch_a}
	\end{subfigure}\hfill
    \begin{subfigure}{0.48\linewidth}
		\includegraphics[width=\linewidth]{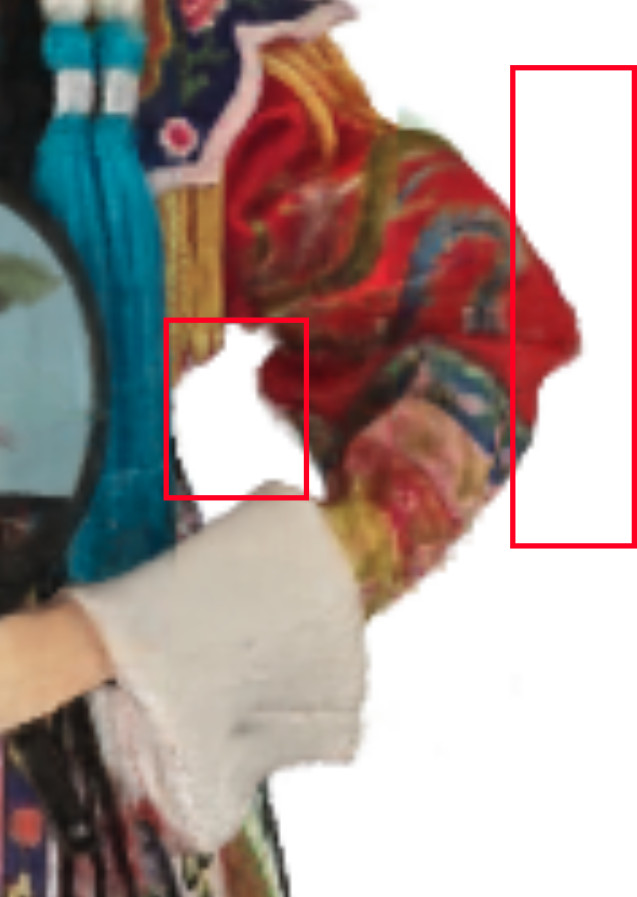}
		\caption{With Distillation}
		\label{fig:from_scratch_b}
	\end{subfigure}
	\vspace{-0.2cm}
	\caption{\textbf{Distillation.} (\subref{fig:from_scratch_a}) Training KiloNeRF from scratch can lead to artifacts in free space. (\subref{fig:from_scratch_b}) Distillation by imitating a pre-trained standard NeRF model mitigates this issue.}
	\label{fig:from_scratch}
\end{figure}

\subsection{Training with Distillation}

Training a KiloNeRF from scratch can lead to artifacts in free space as visualized in Fig.~\ref{fig:from_scratch}. We observed that better results can be obtained by first training an ordinary NeRF model and distilling \cite{hinton2015distill} the knowledge of this teacher into the KiloNeRF model. Here, the KiloNeRF model is trained such that its outputs match the outputs of the teacher model for all possible inputs. More specifically, for each of our networks, we randomly sample a batch of 3D points inside the 3D grid cell that corresponds to the respective network. These batches are augmented with viewing directions which are drawn randomly from the unit sphere. We query both the student and the teacher in order to obtain the respective densities $\sigma$ and color values $\bc$. The obtained $\sigma$-values are converted to $\alpha$-values according to Eq.~\ref{eq:density_to_alpha}, to put less emphasis on small differences between big density values.

We optimize the student's parameters using an $L_2$ loss between the $\alpha$-values and color values predicted by the student and those obtained from the teacher.
Note that for the distillation step we are neither performing volumetric rendering nor are we utilizing the training images. Therefore, distillation alone would imply that KiloNeRF's rendering quality is upper bounded by the teacher method's rendering quality. Hence, in a final step KiloNeRF is fine-tuned on the training images with the photometric loss from Eq.~\ref{eq:photometric_loss}. From this perspective, distillation can be seen as a means to provide powerful weight initialization to KiloNeRF.

\boldparagraph{Regularization}
The original NeRF architecture uses a small fraction of its total capacity to model the dependency of the color on the viewing direction as can be observed in Fig.~\ref{fig:arch} (left). Zhang \etal \cite{zhang2020nerfpp} postulate that this choice encourages that color is a simple function of the viewing direction. Further, they experimentally demonstrate how this inductive bias is crucial for avoiding artifacts in empty space \cite{zhang2020nerfpp}. Our first attempt was to replicate this strategy in KiloNeRF, \ie, we reduced the number of output features of the last hidden layer to get a similar regulatory effect. However, this lead to an overall loss in visual quality due to our small network width. We therefore instead apply $L_2$ regularization to the weights and biases of the last two layers of the network which are responsible for view-dependent modeling of the color. This strategy allows us to impose the same inductive bias as in NeRF without sacrificing visual quality. We refer to \figref{fig:no_l2} in our experiments for an illustration.

\subsection{Sampling}

A maximum of $K$ equidistant points are sampled along the ray. To reduce the number of network queries, we make use of empty space skipping (ESS) and early ray termination (ERT). Consequently, the number of sampled points per ray is variable and the hyperparameter $K$ effectively controls the distance $\delta$ between sampled points. Thanks to the combination of ESS and ERT, only points in the vicinity of the first surface that a ray intersects are evaluated, given that the intersection is with an opaque object. We do not make use of the hierarchical sampling scheme introduced in NeRF, which also enables dense sampling in occupied space, but is costlier than ESS. Similar to NeRF \cite{mildenhall2020nerf}, we use stratified sampling to avoid bias.

\boldparagraph{Empty Space Skipping}
For empty space skipping, we instantiate a second uniform grid with a higher grid resolution in the given AABB. Each cell in this grid is populated with a binary value which indicates whether the scene has any content within that cell. This occupancy grid can be used to avoid querying the network in empty space \cite{beyer2015volsurv}. Only if the cell, in which a sample point $\bx$ lies, is occupied, the network gets evaluated at $\bx$. We extract the occupancy grid from the trained teacher and use it for more efficient fine-tuning and inference. To populate the occupancy grid, we sample densities on a $3\times3\times3$ subgrid from the teacher model for each cell of the occupancy grid. A cell is marked as occupied, if any of the evaluated densities is above a threshold $\tau$. No dataset-specific tuning of $\tau$ was necessary.

\boldparagraph{Early Ray Termination}
If the transmittance value $T_i$ becomes close to 0 during volumetric rendering, evaluation of subsequent sample points $\bx_{i+1}, \bx_{i+2}, \ldots$ is unnecessary, since the contribution of corresponding color values $\bc_{i+1}, \bc_{i+2}, \ldots$ to the final pixel value becomes vanishingly small. Therefore, once transmittance $T_i$ falls below a threshold $\epsilon$, sampling further points along the ray can be avoided. ERT is only used during inference and is implemented by grouping network queries according to their distance from the camera, \ie, points close to the camera are evaluated first and then successively more distant points are queried for those rays that are not terminated yet.

\subsection{Implementation}

Our prototype implementation is based on PyTorch \cite{paszke2019pytorch}, MAGMA \cite{abdelfattah_2017_vbatched} and Thrust. Additionally, we developed custom CUDA kernels for sampling, empty space skipping, early ray termination, positional encoding, network evaluation and alpha blending. Crucial for high performance is the proper handling of the simultaneous query of thousands of networks. To illustrate this, let us consider a single linear layer of a fully connected network which requires the computation of a single matrix multiplication. More specifically, a $B \times I$ matrix is multiplied with an $I \times O$ matrix, where $B$ is the batch size and $I$/$O$ is the number of input/output features. In KiloNeRF, for each network a separate matrix multiplication is required. Assuming $N$ networks in total, we need to multiply a $B_i \times I$ matrix with an $I \times O$ matrix for each $i \in {1, \ldots, N}$. Since $B_i$ varies per network, traditional batched matrix multiplications cannot be used. To tackle this problem, we exploit the HPC library MAGMA which provides a CUDA routine for this situation \cite{abdelfattah_2017_vbatched}. For inference only we use a self-developed routine that fuses the entire network evaluation into a single CUDA kernel. Further, it is necessary to order the input batch such that subsequent inputs are processed by the same network. This step benefits largely from the prior reduction in number of samples caused by the interplay of ESS and ERT. More details can be found in the supplementary.

\section{Experimental Evaluation}

\begin{table*}[]
	\centering
	\begin{tabular}{@{}llrrrr@{}}
		\toprule
		&          & BlendedMVS & Synthetic-NeRF & Synthetic-NSVF & Tanks \& Temples \\
		Resolution & & $768 \times 576$ & $800 \times 800$ & $800 \times 800$  & $1920 \times 1080$ \\ \midrule
		PSNR $\uparrow$                         & NeRF     & 27.29      & 31.01           & 31.55           & 28.32            \\
		& NSVF     & 26.90      & \textbf{31.74}           & \textbf{35.13}           & 28.40            \\
		& KiloNeRF & \textbf{27.39}      & 31.00           & 33.37           & \textbf{28.41}            \\ \midrule
		SSIM $\uparrow$                         & NeRF     & 0.91       & \textbf{0.95}            & 0.95            & 0.90             \\
		& NSVF     & 0.90       & \textbf{0.95}            & \textbf{0.98}            & 0.90             \\
		& KiloNeRF & \textbf{0.92}       & \textbf{0.95}            & 0.97            & \textbf{0.91}             \\ \midrule
		LPIPS $\downarrow$                      & NeRF     & 0.07       & 0.08            & 0.04            & 0.11             \\
		& NSVF     & 0.11       & 0.05            & \textbf{0.01}            & 0.15             \\
		& KiloNeRF & \textbf{0.06}       & \textbf{0.03}            & 0.02            & \textbf{0.09}             \\ \midrule
		Render time (milliseconds) $\downarrow$ & NeRF     & 37266     & 56185          & 56185          & 182671          \\
		& NSVF     & 4398      & 4344           & 10497           & 15697           \\
		& KiloNeRF & \textbf{30}        & \textbf{26}              & \textbf{26}              & \textbf{91}              \\ \midrule
		Speedup over NeRF $\uparrow$             & NSVF     & 8         & 13              & 5              & 12               \\
		& KiloNeRF & \textbf{1258}        & \textbf{2165}             & \textbf{2167}             &\textbf{2002}              \\ \bottomrule
	\end{tabular}
	\caption{\textbf{Quantitative Results.} KiloNeRF achieves similar quality scores as the baselines while being significantly faster. }
	\label{tab:results}
\end{table*}

\subsection{Setup}

\boldparagraph{Datasets}
Both NSVF \cite{liu2020nsvf} and KiloNeRF assume scenes to be bounded. We can therefore directly use the four datasets provided by NSVF, together with the given bounding boxes. Consequently, KiloNeRF is tested on both synthetic datasets (Synthetic-NeRF, Synthetic-NSVF) as well as datasets comprising real scenes (NSVF's variants of BlendedMVS \cite{Yao2020CVPR} and Tanks\&Temples \cite{Knapitsch2017SIGGRAPH}). In total, we evaluate on 25 scenes.

\boldparagraph{Baselines}
We compare KiloNeRF to NeRF \cite{mildenhall2020nerf} and NSVF \cite{liu2020nsvf}. Note that we compare to the original NeRF model which adopts the hierarchical sampling strategy despite it is not used for our teacher. NSVF is particularly interesting as a baseline, because NSVF also makes use of ESS and ERT. Hence speedups over NSVF can be largely attributed to the smaller network size used in KiloNeRF. For comparisons to other NVS techniques \cite{lombardi2019nv, Sitzmann2019NIPS} we refer the reader to \cite{liu2020nsvf}.

\boldparagraph{Hyperparameters}
For each scene, we choose the network grid resolution such that the largest dimension equals $16$ and the other dimensions are chosen such that the resulting cells are cubic, hence the maximally possible resolution is $16\times16\times16=4096$. To arrive at the occupancy grid's resolution we multiply the network grid resolution by 16, which implies that the occupancy grid resolution is upper bounded by $256\times256\times256$ cells. We find that setting the threshold $\tau$ for occupancy grid extraction to $10$ works well on all datasets. We use the Adam optimizer with a learning rate of $5\mathrm{e}{-4}$, a batch size of 8192 pixels and the same learning rate decay schedule as NeRF. For $L_2$ regularization we use a weight of $1\mathrm{e}{-6}$. For early ray termination, we set the threshold to $\epsilon = 0.01$. As found in NSVF, early ray termination with this choice of $\epsilon$ does not lead to any loss of quality \cite{liu2020nsvf}. Further, using the same value for $\epsilon$ as NSVF increases comparability. The hyperparameter $K$, which indirectly controls the distance between sample points, is set to $384$. We use the same number of frequencies for the positional encoding as NeRF.

\boldparagraph{Training Schedule}
We train both the NeRF baseline and KiloNeRF for a high number of iterations to find the limits of the representation capabilities of the respective architectures. In fact, we found that on some scenes NeRF takes up to 600k iterations to converge. When comparing our qualitative and quantitative results to those reported by Liu \etal \cite{liu2020nsvf}, one might notice that we obtained much better results for the NeRF baseline, which might be explained by our longer training schedule.
Training the NeRF baseline on an NVIDIA GTX 1080 Ti takes approximately 2 days. For our teacher, we train another NeRF model but without hierarchical sampling for 600k iterations as NeRF's coarse model is too inaccurate for distillation and NeRF's fine model is biased towards surfaces. This model is distilled for 150k iterations into a KiloNeRF model, which takes up to 8 hours. Finally, the KiloNeRF model is fine-tuned for 1000k iterations, which requires only 17 hours thanks to ESS and the reduced size of our MLPs.

\boldparagraph{Measurements}
Our test system consists of an NVIDIA GTX 1080 Ti consumer GPU, an Intel i7-3770k CPU and 32GB of RAM. For each scene, inference time is measured by computing the average time over all test images.

\subsection{Results}

\begin{figure*}
	\centering
	\includegraphics[width=0.99\linewidth]{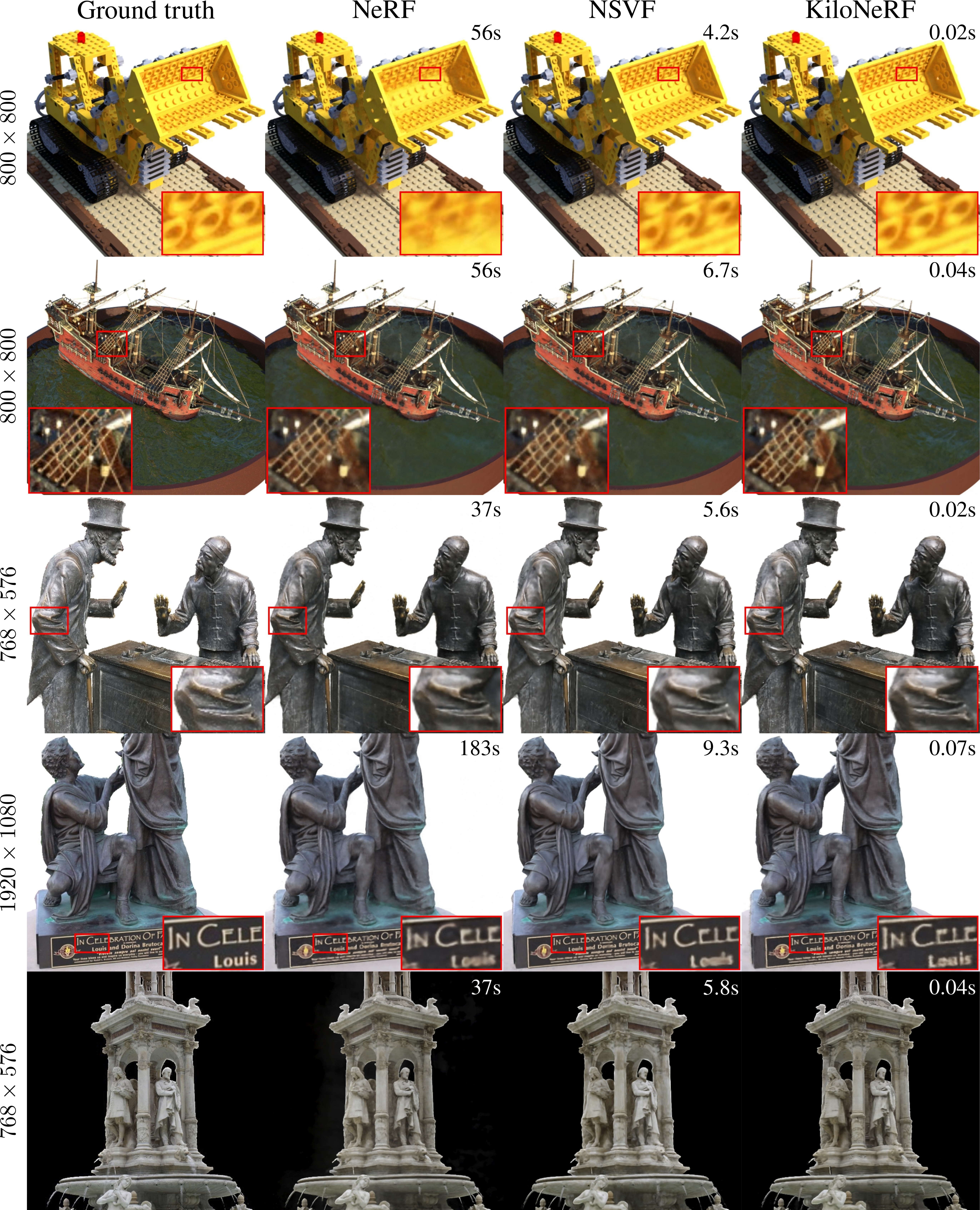}
	\caption{\textbf{Qualitative Comparison.} Novel views synthesized by NeRF, NSVF and KiloNeRF. Despite being significantly faster, KiloNeRF attains the visual quality of the baselines. The numbers in the top-right corner correspond to the average render time of the respective technique on that scene. The rendered image resolution (in pixels) is specified on the left.}
	\label{fig:novel_views}
\end{figure*}

Table~\ref{tab:results} shows our main results. KiloNeRF attains similar visual quality scores as the baselines NeRF and NSVF. In terms of the perceptual metric LPIPS \cite{zhang2018perceptual}, KiloNeRF even slightly outperforms the other two methods on three out of four datasets. However, on all scenes, KiloNeRF renders three orders of magnitude faster than NeRF and two orders of magnitude faster than NSVF. In \figref{fig:novel_views}, we show novel view synthesis results for all three methods. More qualitative results can be found in the supplementary.

\subsection{Ablations}
\begin{figure*}
	\includegraphics[width=\linewidth]{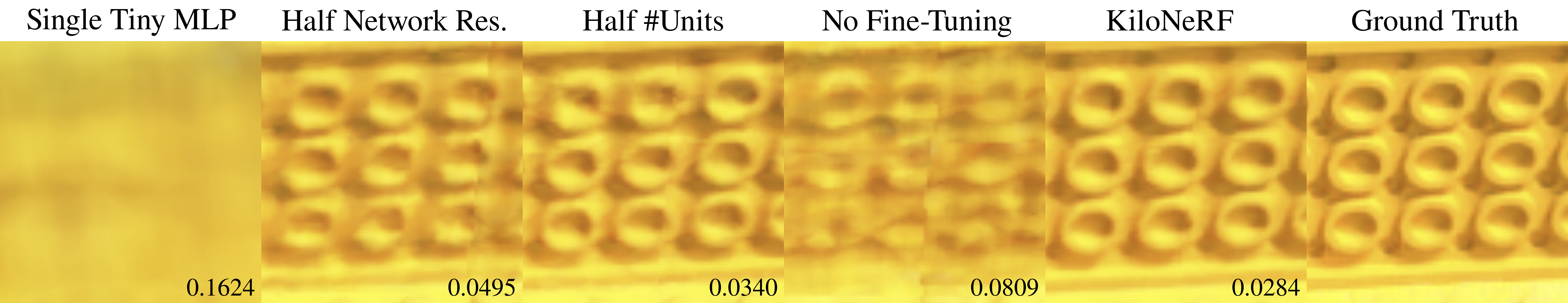}
	\caption{\textbf{Ablation Study.} Closeups of KiloNeRF on the Lego bulldozer scene, varying different parameters of the model. The numbers in the bottom-right corner correspond to perceptual similarity (LPIPS) \wrt the ground truth, lower is better.}
	\label{fig:lego_ablations}
\end{figure*}

\begin{figure}	
	 \begin{subfigure}{0.33\linewidth}
		\includegraphics[width=\linewidth]{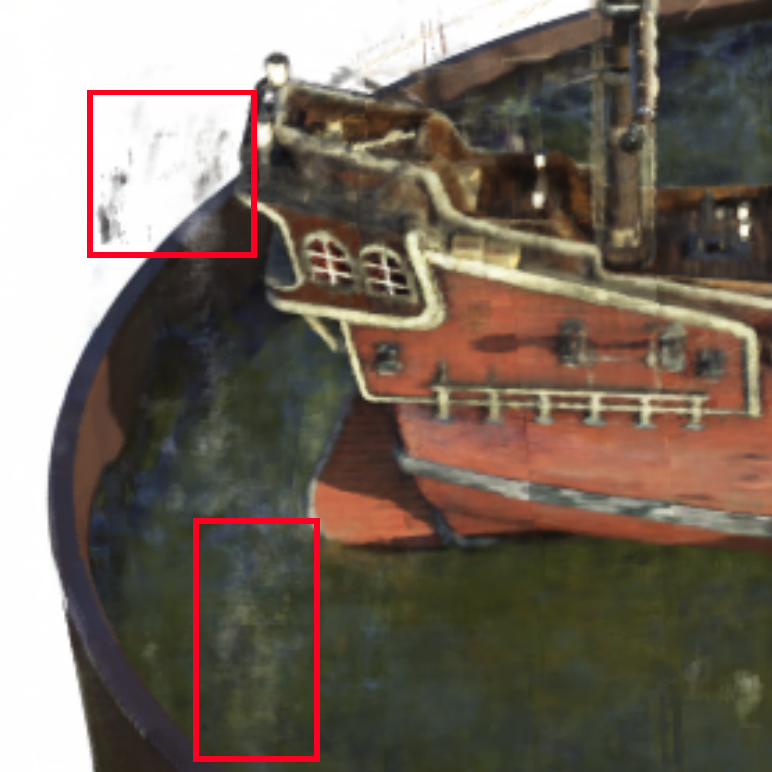}
		\caption{No Regularization}
		\label{fig:no_l2_a}
	\end{subfigure}\hfill
	 \begin{subfigure}{0.33\linewidth}
		\includegraphics[width=\linewidth]{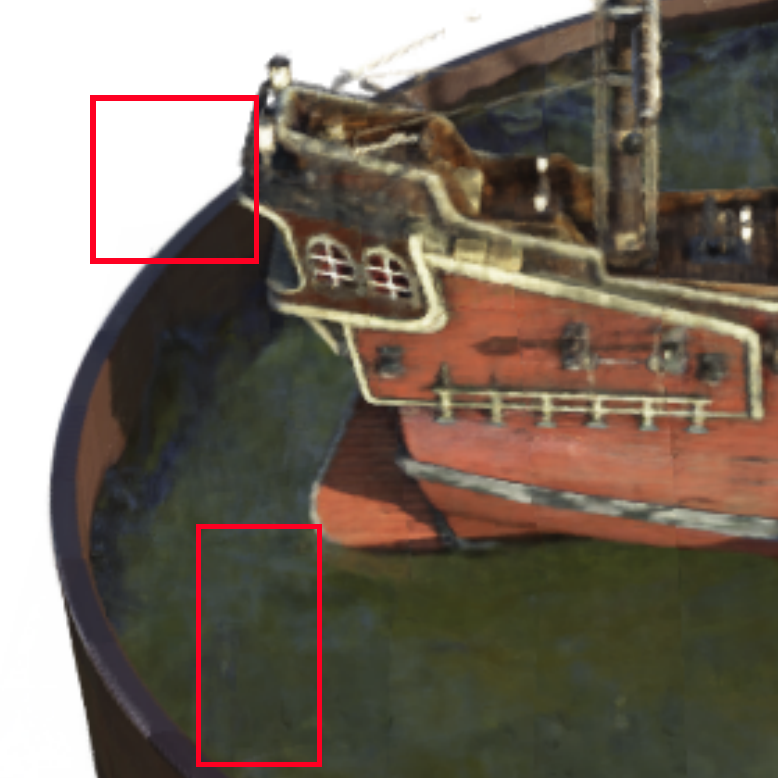}
		\caption{Regularization}
		\label{fig:no_l2_b}
	\end{subfigure}\hfill
	 \begin{subfigure}{0.33\linewidth}
		\includegraphics[width=\linewidth]{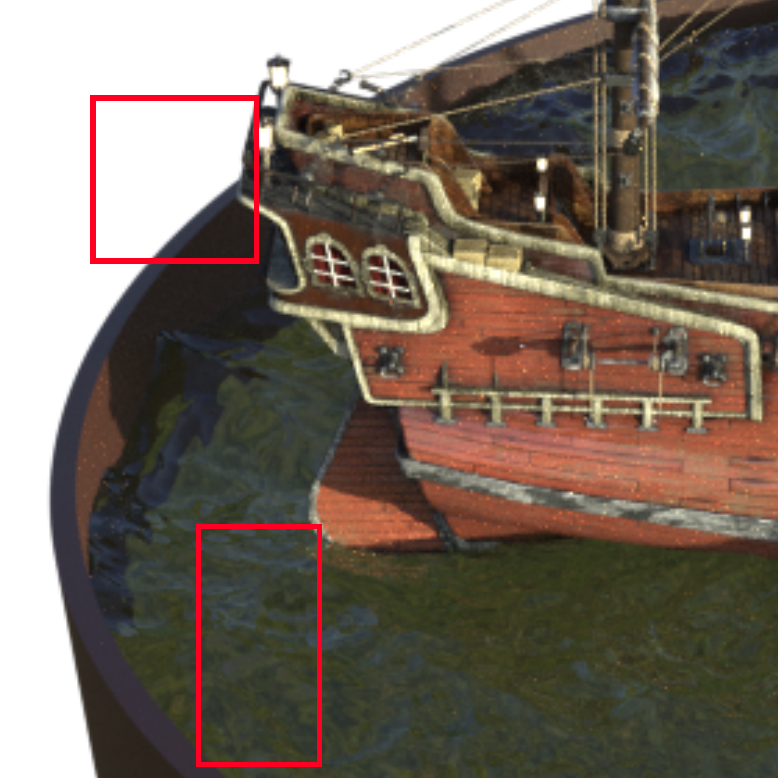}
		\caption{Ground Truth}
		\label{fig:no_l2_c}
	\end{subfigure}\hfill
	\caption{\textbf{Regularization.} Without weight regularization, visible artifacts in free space regions emerge in the rendered images as the view-dependent part of the MLP has too much capacity. Adding $L_2$ regularization alleviates this problem.}
	\label{fig:no_l2}
\end{figure}

\begin{table}[]
	\centering
	\begin{tabular}{@{}lrr@{}}
		\toprule
		Method           & Render time $\downarrow$ & Speedup $\uparrow$ \\ \midrule
		NeRF             & 56185 ms                          & --                  \\
		NeRF + ESS + ERT & 788 ms                            & 71                 \\
		KiloNeRF         & \textbf{22} ms                    & \textbf{2548}       \\ \bottomrule
	\end{tabular}
	\caption{\textbf{Speedup Breakdown.} The original NeRF model combined with KiloNeRF's implementation of ESS and ERT is compared against the full KiloNeRF technique.}
	\label{tab:speed_ablation}
\end{table}

In \figref{fig:lego_ablations}, we conduct an ablation study on KiloNeRF using the Lego bulldozer scene as an example.

\boldparagraph{Single Tiny MLP}
For this ablation, KiloNeRF's network hyperparameters (number of layers/hidden units) are used, but only a single network gets instantiated. Unsurprisingly, quality suffers dramatically since a single MLP with only 6k parameters cannot accurately represent the entire scene.

\boldparagraph{Half Network Resolution}
We test KiloNeRF with half the network grid size ($5\times8\times5$ instead of $10\times16\times10$ for this particular scene) to verify whether the chosen resolution is not unnecessarily high. We found that reducing the grid resolution leads to a decline in quality which demonstrates that a fine grid of tiny networks is indeed beneficial.

\boldparagraph{Half \#Units}
We also tested whether it is possible to further reduce the number of hidden units. Again, we observe a decrease in image quality, although the loss of quality is not as pronounced as when decreasing grid resolution. A possible explanation might be that halving the network resolution decreases the number of parameters by a factor of 8, while halving the number of hidden units decreases the parameter count only by approximately a factor of 4.

\boldparagraph{No Fine-Tuning}
For this experiment, the ultimate fine-tuning phase is omitted. Our results show that fine-tuning is crucial for achieving high quality and that the prior distillation phase only provides a good parameter initialization.

\boldparagraph{No Distillation}
Fig.~\ref{fig:from_scratch} compares training KiloNeRF from scratch against the proposed training pipeline. Both alternatives yield the same level of detail, but only when using distillation artifacts in empty space are avoided.

\boldparagraph{No Weight Regularization}
Fig.~\ref{fig:no_l2} demonstrates that artifacts in free space can also emerge when $L_2$ regularization on the last two layers of the network is omitted. Consequently, both distillation and $L_2$ regularization play important roles in avoiding these artifacts.

\boldparagraph{ESS and ERT}
Finally, we are interested in determining to which extent the combination of ESS and ERT and the reduction of network size contribute to the final speedup. Towards this goal, we run the original-size NeRF with our optimized ESS/ERT implementation and compare to the full KiloNeRF technique on the Lego bulldozer scene. As can be deduced from Table~\ref{tab:speed_ablation}, the reduction in network size contributes significantly to KiloNeRF's overall speedup.

\section{Discussion and Future Work}

While KiloNeRF is able to render medium resolution images ($800 \times 800$) at interactive frame rates, the speedups are not sufficient yet for real-time rendering of full HD images. Further speedups might be possible by scaling our approach to a higher number of smaller networks. Na\"{i}vely doing this, however, would lead to a higher storage impact. This might be mitigated by using memory-efficient data structures \cite{laine2010svo, dai_2015_regress} that allow networks to be exclusively instantiated in the vicinity of surfaces.

Furthermore, in this work, we mainly focused on making network queries faster by using many small MLPs. In contrast, the concurrently developed techniques AutoInt \cite{lindell2020autoint} and DONeRF \cite{Neff2021donerf} speed up inference by reducing the number of required samples along the ray. Therefore, the combination of KiloNeRF with either of these techniques constitutes a promising solution for achieving real-time rendering of high-resolution images in the near future.

\boldparagraph{Limitations}
NSVF \cite{liu2020nsvf} and KiloNeRF share the assumption of a bounded scene which is a limitation that should be addressed in future work. Representing unbounded scenes requires a higher number of networks, leading to larger memory consumption. Again, efficient data structures could help with addressing this issue and therefore allow for scaling to larger (\eg, outdoor) scenes. On the other hand, the scenes considered in this paper are encoded with less than 100 MB of memory, respectively. This shows that -- at least for medium-sized scenes -- storage is not an issue.

\section{Conclusion}
In this paper, we demonstrated that real-time rendering of NeRFs can be achieved by spatially decomposing the scene into a regular grid and assigning a small-capacity network to each grid cell. Together with the other benefits inherited from NeRF -- excellent render quality and a low storage impact -- we believe that this constitutes an important step towards practical NVS. Moreover, the presented acceleration strategy might also apply more broadly to other methods relying on neural function representations including implicit surface models.

\boldparagraph{Acknowledgements}
This work was supported by an NVIDIA research gift, the ERC Starting Grant LEGO-3D (850533) and the DFG EXC number 2064/1 - project number 390727645. Songyou Peng is supported by the Max Planck ETH Center for Learning Systems.  Yiyi Liao is supported by the BMBF Tübingen AI Center, FKZ: 01IS18039A.

{\small
 \bibliographystyle{ieee_fullname}
 \bibliography{bibliography_long,bibliography_custom,bibliography}
}

\end{document}